\documentclass[runningheads]{llncs}
\usepackage{graphicx}
\usepackage[misc]{ifsym}
\usepackage{tikz}
\usepackage{comment}
\usepackage{amsmath,amssymb} % define this before the line numbering.
\usepackage{color}

\usepackage[accsupp]{axessibility}  % Improves PDF readability for those with disabilities.

\usepackage{algorithm}
\usepackage{mathrsfs}
\usepackage{booktabs}
\usepackage[pagebackref,breaklinks,colorlinks]{hyperref}
\usepackage{algpseudocode}
\usepackage{multirow}
\usepackage{caption}
\usepackage{subcaption}
\usepackage{diagbox}
\usepackage{bm}
\usepackage{orcidlink}

\newcommand\blfootnote[1]{%
\begingroup
\renewcommand\thefootnote{}\footnote{#1}%
\addtocounter{footnote}{-1}%
\endgroup
}

\usepackage{xcolor,colortbl}
\definecolor{Gray}{gray}{0.90}
\definecolor{LightCyan}{rgb}{0.88,1,1}
\newcolumntype{a}{>{\columncolor{Gray}}c}

\usepackage[pagebackref,breaklinks,colorlinks]{hyperref}

\usepackage[capitalize]{cleveref}
\crefname{section}{Sec.}{Secs.}
\Crefname{section}{Section}{Sections}
\Crefname{table}{Table}{Tables}
\crefname{table}{Tab.}{Tabs.}

\usepackage{xspace}
\DeclareRobustCommand\onedot{\futurelet\@ }
\def\@onedot{\ifx\@let@token.\else.\null\fi\xspace}
\def\eg{\emph{e.g}\onedot} 
\def\ie{\emph{i.e}\onedot}

\def\etal{\emph{et al}\onedot}

\begin{document}
% \linenumbers
\pagestyle{headings}
\mainmatter
\def\ECCVSubNumber{2033}  % Insert your submission number here

\title{PoseTrans: A Simple Yet Effective Pose Transformation Augmentation for Human Pose Estimation} % Replace with your title

% \begin{comment}
% \titlerunning{ECCV-22 submission ID \ECCVSubNumber} 
% \authorrunning{ECCV-22 submission ID \ECCVSubNumber} 
% \author{Anonymous ECCV submission}
% \institute{Paper ID \ECCVSubNumber}
% \end{comment}
%******************

% \begin{comment}
\titlerunning{PoseTrans}
\author{
    Wentao Jiang\inst{1,2}\orcidlink{0000-0002-4894-5703} \and
    Sheng Jin\inst{3,2}\orcidlink{0000-0001-5736-7434} \and
  Wentao Liu\inst{2,4}\orcidlink{0000-0001-6587-9878} \and
  Chen Qian\inst{2}\orcidlink{0000-0002-8761-5563}  \\  
  Ping Luo\inst{3}\orcidlink{0000-0002-6685-7950} \and  
  Si Liu\inst{1,5}\textsuperscript{\Letter}
   \\[.21cm]
  $^{1}$Institute of Artificial Intelligence, Beihang University \\
  $^{2}$SenseTime Research and Tetras.AI \\
  $^{3}$The University of Hong Kong \quad $^{4}$Shanghai AI Laboratory \\
  $^{5}$State Key Lab. of VR Technology and Systems, SCSE, Beihang University \\
  \tt\small \{jiangwentao, liusi\}@buaa.edu.cn \quad js20@connect.hku.hk \\
  \tt\small \{liuwentao, qianchen\}@sensetime.com \quad
   pluo@cs.hku.hk
  }
\authorrunning{W. Jiang et al.}
\institute{}
% \end{comment}

%******************
\maketitle

\begin{abstract}
  Human pose estimation aims to accurately estimate a wide variety of human poses. However, existing datasets often follow a long-tailed distribution that unusual poses only occupy a small portion, which further leads to the lack of diversity of rare poses. These issues result in the inferior generalization ability of current pose estimators. In this paper, we present a simple yet effective data augmentation method, termed Pose Transformation (PoseTrans), to alleviate the aforementioned problems. Specifically, we propose Pose Transformation Module (PTM) to create new training samples that have diverse poses and adopt a pose discriminator to ensure the plausibility of the augmented poses. Besides, we propose Pose Clustering Module (PCM) to measure the pose rarity and select the ``rarest'' poses to help balance the long-tailed distribution. Extensive experiments on three benchmark datasets demonstrate the effectiveness of our method, especially on rare poses. Also, our method is efficient and simple to implement, which can be easily integrated into the training pipeline of existing pose estimation models. 
  % Codes will be available at \url{https://github.com/wtjiang98/PoseTrans}.


\blfootnote{\Letter~: Corresponding Author.}

\end{abstract}

\section{Introduction}

Human Pose Estimation (HPE) is the task of localizing human body keypoints (also referred to as joints) from an image. It serves as a fundamental technique for numerous applications, including action recognition, pedestrian tracking, and virtual/augmented reality. 
Recently, deep convolutional neural networks (DCNN) \cite{toshev2014deeppose,newell2016stacked,newell2017associative} have achieved drastic improvements on standard benchmark datasets.
To fully exploit the power of DCNN, a large number of training data is indispensable for obtaining satisfactory performance in human pose estimation.

\begin{figure}[!t]
   \centering
   \includegraphics[width=0.9\linewidth]{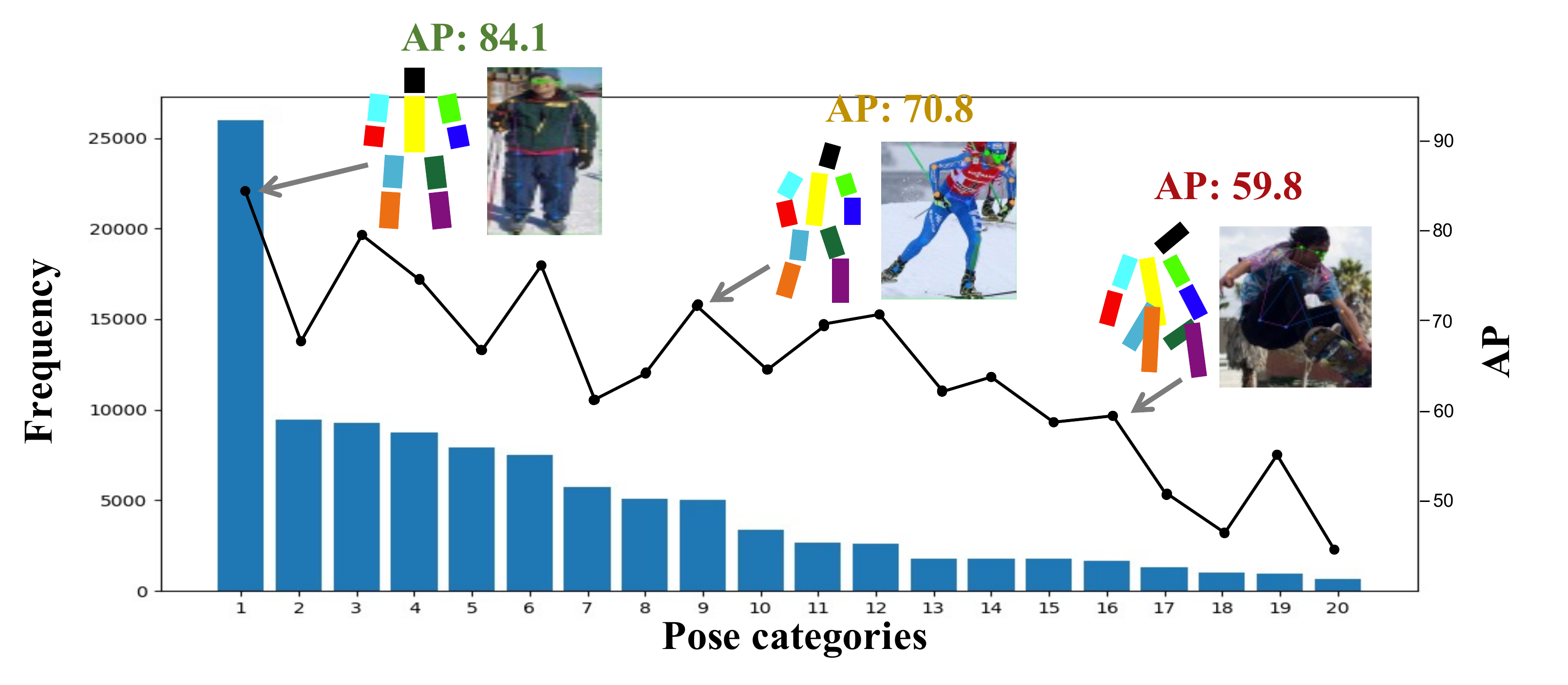}
   \caption{
       We cluster the poses in the MS-COCO dataset into 20 categories and evaluate the AP with a pre-trained HRNet model~\cite{wang2020deep}. The top-1 category has more than 25000 samples and high precision, while nearly half of the categories have less than 2000 samples and relatively low precision.
   }
   \label{fig1}
\end{figure}

However, existing human pose estimation datasets do not uniformly represent all possible human poses in real life. We take MS-COCO dataset~\cite{lin2014microsoft} as an example to analyze the distribution of the human poses, as shown in Fig. \ref{fig1}. We normalize the poses and cluster them into 20 categories. We observe that it follows a long-tailed distribution, with a few common pose categories (\eg standing and walking) occupying a large portion of the dataset and unusual posture types (\eg squatting and jumping) possessing a smaller portion.
We also find that although current state-of-the-art data-driven methods achieve good performance on common poses, however, they still suffer performance degradation on some unusual poses, since the long-tailed categories have neither enough training samples nor enough diversity.

Due to the high cost of collecting and annotating examples with rare poses, a feasible way to tackle this problem is data augmentation. 
Previous methods augment the human pose mainly by global image-level transformations \cite{peng2018jointly,chen2017adversarial,newell2016stacked,xiao2018simple,wang2021human} (\eg scaling and rotating) or local object-level transformations \cite{bin2020adversarial,peng2018jointly,fang2019instaboost} (\eg copy-paste and occluding).
Since these methods fail to increase the diversity of poses and alleviate the long-tailed distribution, they contribute little to recognizing diverse rare poses.

In this paper, we propose a simple yet effective data augmentation approach, termed Pose Transformation (PoseTrans), to tackle the aforementioned challenges. PoseTrans consists of a Pose Transformation Module (PTM) with a pose discriminator, and a Pose Clustering Module (PCM).
During training, PTM applies affine transformations to the original pose of the training sample and generates a pool of diverse new poses.
The pre-trained pose discriminator is adopted to evaluate the plausibility of generated samples and then filter out unnatural samples.
PCM is based on the Gaussian Mixture Model (GMM), which normalizes and clusters the human poses in the dataset. The rare types of poses are represented by the Gaussian components that have small weights. 
PCM evaluates the components' density for each candidate pose and selects the ``rarest'' one (\ie which has the minimal weighted sum of components' density) as the final augmented training sample.
By transforming the existing poses, PoseTrans helps generate diverse, plausible poses by PTM and alleviate the long-tail distribution problem by PCM.
We also design a metric that focuses on rare poses called balanced AP/AR and observe more performance gain on this metric.
Our method is simple to implement and can be easily integrated into the training pipeline of existing pose estimation models. 

We summarize our contributions as follows:

\begin{enumerate}
   
   \item[$\bullet$] We present a simple yet effective data augmentation method, termed PoseTrans. To tackle the problem of limited diversity of unusual human poses, we propose a novel Pose Transformation Module (PTM) with a pose discriminator to generate new training samples with diverse and plausible poses.
   
   \item[$\bullet$] We propose Pose Clustering Module (PCM) to measure the pose rarity and select rare poses for data augmentation, which helps to balance the long-tailed distribution of the training set.
   
   \item[$\bullet$] Extensive experiments on various pose estimation datasets show that PoseTrans consistently improves the performance of various state-of-the-art pose estimators, especially on rare poses.
\end{enumerate}

\section{Related Works}

\subsection{2D Human Pose Estimation}
In recent years, 2D human pose estimation has shown remarkable performance advancement. DeepPose \cite{toshev2014deeppose} first applied deep neural networks to human pose estimation by directly regressing the 2D coordinates of key points from the input image.
Since then, deep learning-based methods started to dominate this area.
Recent multi-person human pose estimation approaches can be divided into bottom-up and top-down approaches. Bottom-up approaches~\cite{Insafutdinov2016DeeperCut,cao2017realtime,papandreou2018personlab,newell2017associative,jin2020differentiable,cheng2020higherhrnet,kreiss2019pifpaf,jin2019multi} first detect all the key points of every person in images and then group them into individuals. Top-down methods~\cite{he2017mask,chen2018cascaded,xiao2018simple,sun2019deep} first detect the bounding boxes and then predict the human body key points in each box.

Recent works mainly focus on designing powerful network architectures to improve the performance of pose estimation~\cite{newell2016stacked,xiao2018simple,sun2019deep,chen2018cascaded,jin2020whole,xu2021vipnas,zeng2022not}. However, current state-of-the-art models often suffer performance drops on rare poses due to the long-tailed distribution problem in human pose data. In this work, we focus on tackling this important but ignored problem.
Standing on the shoulder of the well-designed network structure, we propose a novel data augmentation method to generate diverse rare poses.

\subsection{Data Augmentation}

Data augmentation has been widely utilized to improve the model generalization ability. For image classification, popular augmentation methods include information dropping~\cite{zhong2020random,chen2020gridmask,devries2017improved}, multi-image information mixing~\cite{zhang2017mixup,yun2019cutmix} and automatic augmentation~\cite{cubuk2018autoaugment}.
For human pose estimation, data augmentation mainly focus on global image-level transformations \cite{peng2018jointly,chen2017adversarial,newell2016stacked,xiao2018simple,wang2021human} (\eg scaling, rotating, and flipping) and local object-level transformations \cite{bin2020adversarial,peng2018jointly} (\eg copy-paste, occluding).
These common data augmentation schemes enhance the global translational invariance and robustness in occlusion cases but struggle to improve the immunity to rare poses.
Recently, some augmentation methods \cite{fang2019instaboost,fang2021decaug} propose to perform jitting on instances to increase the generalization of the model, but they do not change either the instance itself or the distribution of instances.
Different from the existing data augmentation strategies, we propose a novel, simple and effective PoseTrans augmentation scheme that directly generates diverse rare poses.

\subsection{Long-tailed Distribution}

In visual recognition, there exists a challenging problem of long-tailed training set distributions, where a small portion of classes have massive training samples while classes in the distribution tail have few samples~\cite{zhang2021deep}. Over-sampling~\cite{chawla2002smote} and re-weighting~\cite{elkan2001foundations} are two popular methods to tackle the problem. The over-sampling method raises the frequency level of the minor classes by repeating the data samples during training. The re-weighting method assigns higher loss weights to these minor classes and thus increases their importance. However, such approaches do not increase the diversity of the data and tend to suffer from over-fitting which leads to a performance drop. Other approaches also include metric learning that enforces inter-class margins \cite{huang2016learning} and meta-learning that learns to regress many-shot model parameters from few-shot model parameters \cite{wang2017learning}, but they are only designed for visual recognition. 
In human pose estimation, we encounter a similar problem. For many human pose estimation datasets~\cite{lin2014microsoft,andriluka14cvpr,li2019crowdpose}, \eg the MS-COCO dataset~\cite{lin2014microsoft}, the distribution of human poses is highly biased, which does not uniformly represent human poses in real life. These dataset biases lead to poor generalization and degraded detection accuracy of these ``long-tailed'' poses. To address the aforementioned issue, we propose a simple yet effective PoseTrans approach to create the needed diverse poses.

\begin{figure*}[!t]
  \centering
  \includegraphics[width=0.95\linewidth]{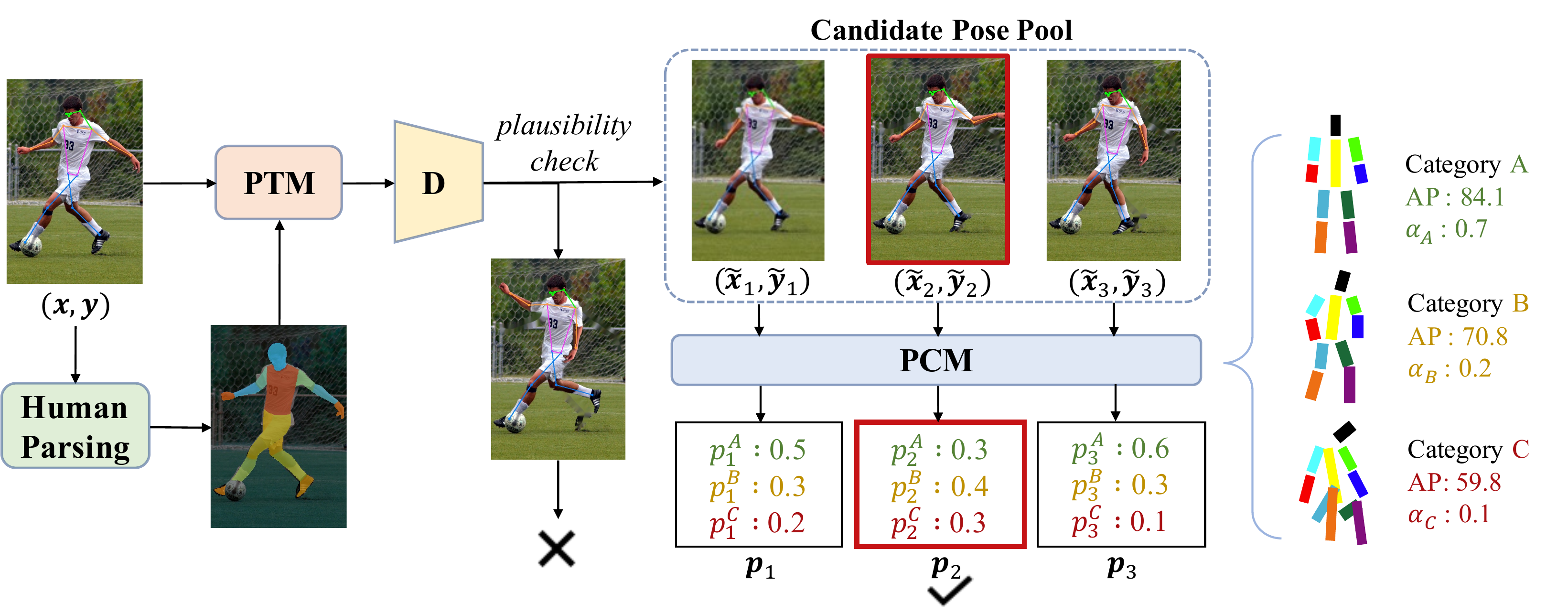}
  \caption{
    Overview of PoseTrans.
    Given a single human image $\bm{x}$ and its keypoint annotations $\bm{y}$, we first segment the human into different parts through human parsing. PTM applies affine transformations on the limbs of the human to construct new poses. A pre-trained pose discriminator is used for the plausibility check. The plausible poses form a candidate pose pool $\{ (\bm{\tilde{x}}_t, \bm{\tilde{y}}_t) \}$, where $t \in \{1, 2, 3\}$ as an example.
     For pose $\bm{\tilde{y}}_t$, PCM predicts $\bm{w}_t$, which is the probability of belonging to each category (3 categories as an example).
     PCM selects the rarest one with the minimal weighted sum of components' density as a new training sample, \ie $w^A_2 \alpha_A + w^B_2 \alpha_B + w^C_2 \alpha_C$.
  }
  \label{framework}
\end{figure*}

\section{Method}
\subsection{Overview}
To increase the diversity of poses and alleviate the long-tailed distribution problem, we propose the Pose Transformation (PoseTrans) to generate new training samples with diverse poses, as shown in Fig.~\ref{framework}.
PoseTrans consists of a Pose Transformation Module (PTM) with a pose discriminator $D$ and a Pose Clustering Module (PCM).
Given a training sample $(\bm{x}, \bm{y})$ consisting of a single human image $\bm{x}$ and its keypoint annotation $\bm{y}$, PTM aims to create a new training sample $(\bm{\tilde{x}}, \bm{\tilde{y}})$ by applying affine transformations on the limbs of the human, where $\bm{x}, \bm{\tilde{x}} \in \mathbb{R}^{H \times W \times 3}$, $\bm{y}, \bm{\tilde{y}} \in \mathbb{R}^{J \times 2}$. $H$, $W$ and $J$ indicate the height, width and the number of keypoints respectively.
To ensure plausibility, we leverage the pose discriminator $D$ to filter out implausible samples.
PoseTrans applies PTM repeatedly until a candidate pose pool with $T$ plausible generated poses is formed.
PCM clusters human poses into $N$ categories and evaluates the probability of belonging to each cluster for generated poses to select the rarest one among the pool as a new training sample.
After each training epoch, we re-fit the PCM using the original training set and all the selected augmented samples.

\subsection{Pose Transformation Module (PTM) and Pose Discriminator}
\label{PT}
\begin{figure}[!t]
  \centering
  \includegraphics[width=0.79\linewidth]{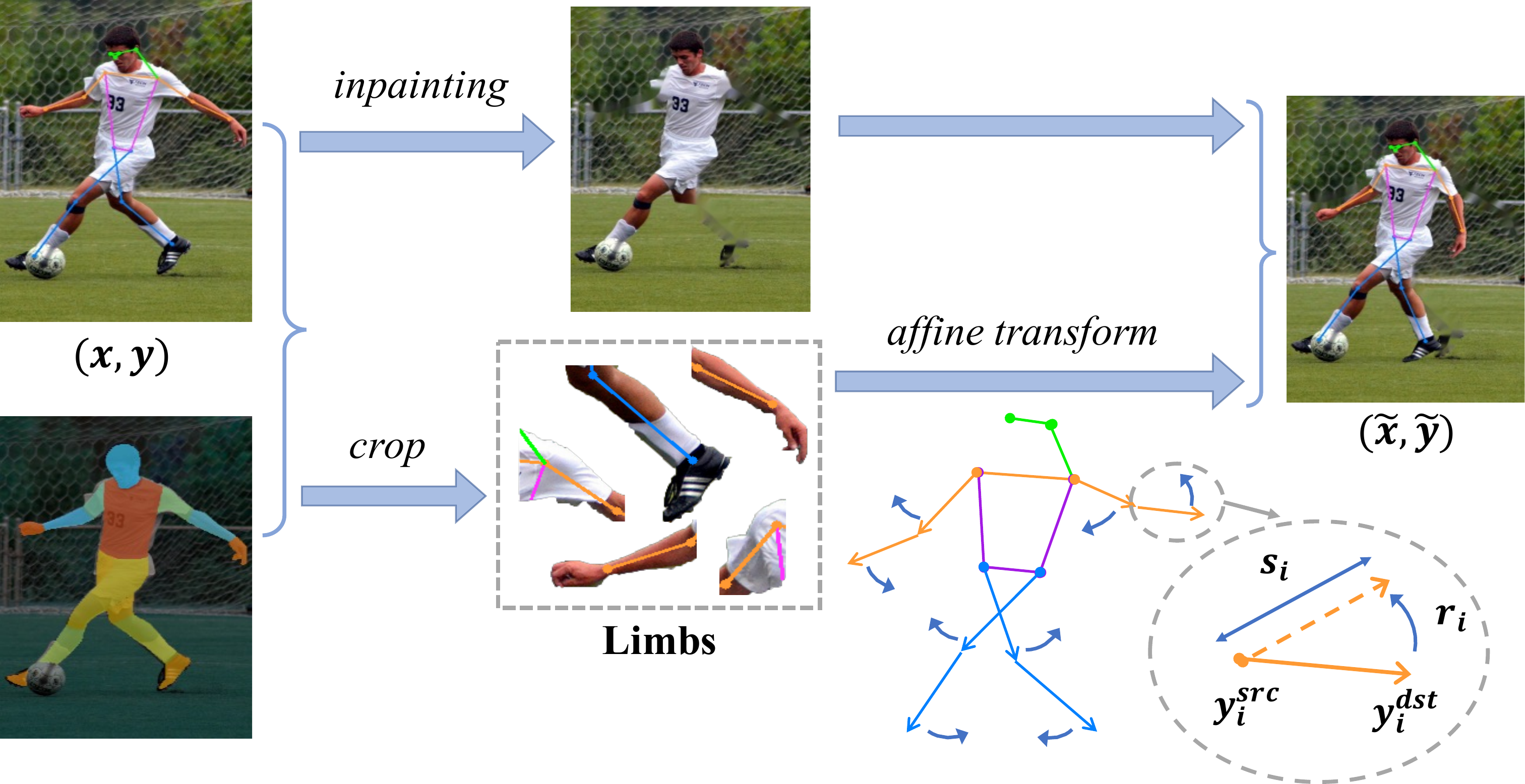}
  \caption{
     By leveraging the human parsing results, we first erase the limbs from $\bm{x}$ and then transform each limb separately with a given probability $p = 0.5$.
     Limbs that do not appear or are obscured will not be transformed.
     The zoom-in view in the bottom right corner indicates the affine transformation with scale $s_i$ and rotation $r_i$ applied on the $i$-th limb (lower arm).
  }
  \label{pt}
\end{figure}

By clustering the human poses in the existing dataset, it can be observed that many clusters only have a few examples.
The lack of training examples of rare poses further leads to the lack of diversity of rare poses, which results in the inferior performance of current data-driven methods on these types of poses.
To tackle this issue, we devise the Pose Transformation Module (PTM) and a pose discriminator to create plausible new poses based on the existing training samples.
The detail of PTM is shown in Fig. \ref{pt}.

\textbf{Modeling the body part movement.}
The body kinematic skeleton is constructed by a pose graph, where the human body is partitioned into several parts, \ie the head, the torso, the left/right arm, and the left/right leg. In this work, we mainly focus on the angular movement of the arms and legs.
Angular movements (flexion and extension) take place at the shoulder, hip, elbow, knee, and wrist. Flexion decreases the angle between the bones (bending of the joint), while extension increases the angle and straightens the joint. 
These body part movements in the image plane can be modeled by applying the affine transformation to a rigid body part segment. In our implementation, the affine transformation is composed of rotation and scaling.

We define the limb as a single rigid body part connecting natural adjacent joints $\bm{y}^{src}$ and $\bm{y}^{dst}$, where $\bm{y}^{src}, \bm{y}^{dst} \in \mathbb{R}^2$ are the coordinates of the source and destination joint respectively. We define $K = 8$ limbs for each instance, including the lower arm, the upper arm, the lower leg, and the upper leg of both sides.

\textbf{Pose transformation.}
With human parsing results obtained through DensePose \cite{alp2018densepose} model, PTM first erases the original limbs in $\bm{x}$ by an efficient inpainting method \cite{bertalmio2001navier}.
After that, each limb is transformed by its affine transformation matrix separately.
To increase the diversity, each limb has a probability of $p=0.5$ to decide whether to transform or not.
The transformed limbs and the inpainted image are composed to form the new augmented image $\bm{\tilde{x}}$. And the pose annotations are also transformed accordingly to get $\bm{\tilde{y}}$.

Specifically, the angular movement of the $i$-th limb can be modeled by the following affine transformation matrix
\begin{small}
\begin{equation}
   \begin{gathered}
      \bm{H}_i = \left[\begin{array}{ccc}
         s_i \cos r_i & - s_i \sin r_i & (1-\cos r_i)c_i^{x} + c_i^y \sin r_i \\
         s_i \sin r_i & s_i \cos r_i & (1-\cos r_i)c_i^{y} - c_i^x \sin r_i \\
         0 & 0 & 1
         \end{array}\right], 
   \end{gathered}
\end{equation}
\end{small}
where $s_i \in \mathbb{R^+}$ and $r_i \in \mathbb{R}$ denote the scale and rotation of the $i$-th limb, $ \bm{y}_i^{src} = \{c_i^{x}, c_i^{y}\}$ is the coordinates of the rotation center of the $i$-th limb.
For the lower arm, the upper arm, the lower leg, and the upper leg, the rotation centers are the elbow, the shoulder, the knee, and the hip respectively.
To ensure the diversity of augmented poses, the scale $s_i$ and rotation $r_i$ parameters in $\bm{H}_i$ are randomly sampled from a normal distribution in the neighboring space of identity transformation $(1, 0)$.
The scale and rotation parameters are also restricted to a certain range in our implementation to ensure that the majority of the randomly generated poses are plausible. Note that, limbs that do not appear in the image or are obscured will not be transformed.

According to the kinematic skeleton hierarchy, the movement of the upper arm/leg will affect that of its lower part. 
Suppose the $j$-th limb is the lower arm/leg and the $k$-th limb is its corresponding upper part. Considering the combined effect, the total movement of the $j$-th limb can be modeled by matrix multiplication, \ie $\bm{H}_{k} \bm{H}_j$.

\textbf{Pose discriminator for the plausibility check.}
Purely generating poses randomly may result in implausible poses that violate the biomechanical structure of the human body.
Some other augmentation methods \cite{li2020cascaded,chen2016synthesizing} rely on pre-defined rules for ensuring plausibility, which however limits the diversity of generated poses.
Inspired by \cite{gong2021poseaug}, we design a pose discriminator $D$ that suits our task to avoid implausible poses that have unnatural joint angles or unreasonable positions in the scene.
For the augmented sample $(\bm{\tilde{x}}_t, \bm{\tilde{y}}_t)$, the discriminator $D$ is trained to predict the plausibility $e_t = D\left(\bm{\tilde{x}}_t, \bm{\tilde{y}}_t\right)$.
We adopt the LS-GAN loss \cite{mao2017least} to train the discriminator before training the pose estimatior:
\begin{equation}
      \begin{aligned}
        \mathcal{L}_{D} = \mathbb{E}\left[\left(D(\bm{x}, \bm{y})-1\right)^{2}\right]+\mathbb{E}\left[D\left(\bm{\tilde{x}}, \bm{\tilde{y}}\right)^{2}\right].
        \end{aligned}
\end{equation}
With the pre-trained discriminator $D$, PoseTrans efficiently filter out the augmented sample whose plausibility is less than a pre-defined threshold $E \in [0, 1]$, and fill the candidate pose pool with samples that are plausible and diverse.

\subsection{Pose Clustering Module (PCM)}
\label{PCM}
After gaining the ability to create new human poses by PTM, we propose the Pose Clustering Module (PCM) to measure the pose rarity and select the needed poses for data augmentation.

\textbf{Fitting the PCM.}
Our PCM is built upon the Gaussian Mixture Model (GMM) with $N$ Gaussian components. 
As a soft clustering method, it predicts the probability of belonging to a certain category.
Before pose clustering, human poses in the training set are first normalized. 
We crop every human instance on the image and re-scale the cropped image into the same height and width ($256 \times 256$). The corresponding keypoint coordinates are also normalized at the same time.
We fit the PCM using the normalized human poses in the training set.
After fitting, given the pose $\bm{y}$, we model $P(\bm{y})$ as:
\begin{equation}
   P(\bm{y}) = \sum_{n=1}^{N} \alpha_{n} \mathcal{N}\left(\bm{y} ; {\mu}_{n}, {\sigma}_{n}\right), 
\end{equation}
where $\alpha_n$ is the weight of the $n$-th Gaussian component, $\mathcal{N}\left(\bm{y} ; {\mu}_{n}, {\sigma}_{n}\right)$ denotes the $n$-th Gaussian distribution with mean ${\mu}_{n}$ and covariance ${\sigma}_{n}$.

\begin{figure}[!t]
  \centering
  \includegraphics[width=0.85\linewidth]{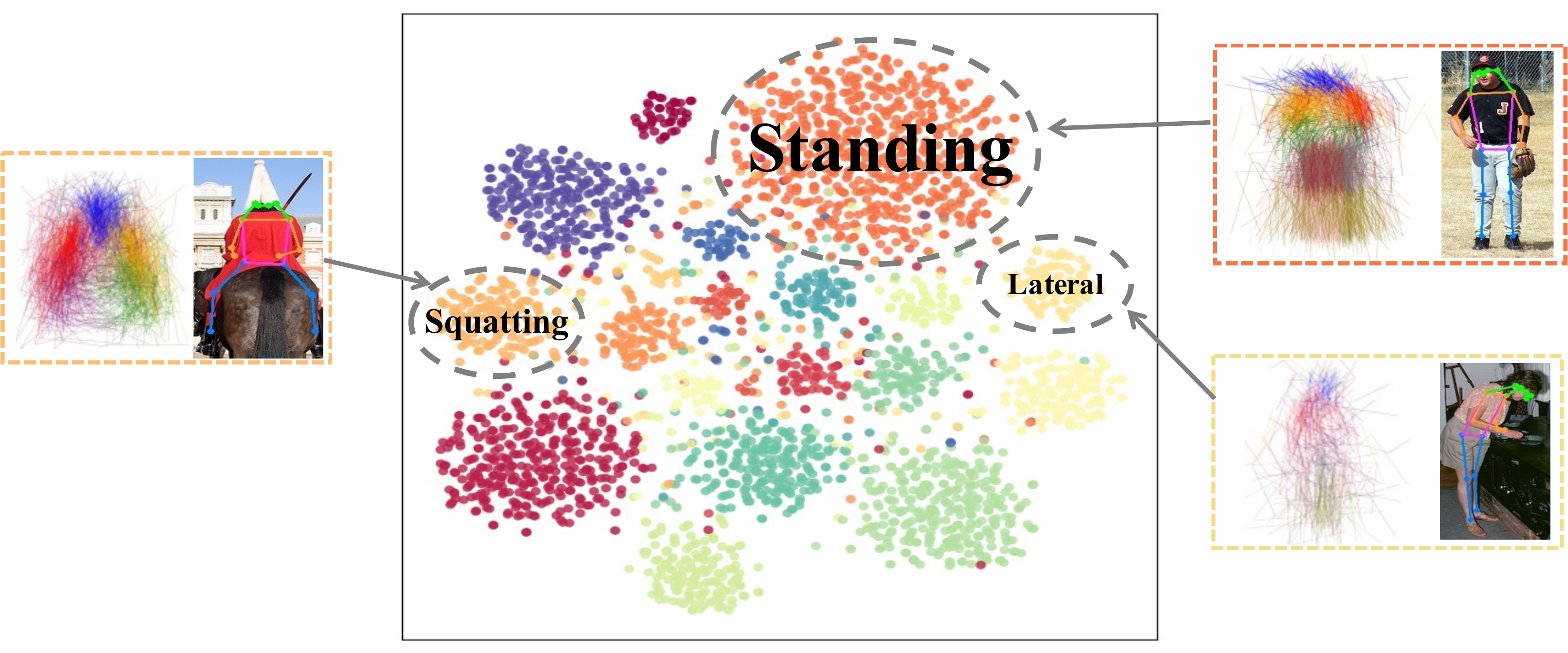}
  \caption{
     The visualization of the clustering results using PCM by t-SNE. Different colored points indicate different clusters. Representative images and mean skeletons for the clusters of standing, squatting, and lateral poses are also visualized.
  }
  \label{tsne}
\end{figure}

By predicting the probability of belonging to each Gaussian component, the human pose is classified as the component with the maximum probability.
We visualize the probability vectors of every example using t-SNE \cite{van2008visualizing}, as shown in Fig. \ref{tsne}.
With PCM, we cluster the human poses into $N$ categories, where
Gaussian components that have small weights (\ie few examples,) indicate the categories of rare poses.
We observe the long-tailed problem that frontal standing accounts for a significant portion while squatting and lateral postures account for small percentages.

\textbf{Pose selection from the candidate pose pool.}
PoseTrans repeats PTM to build a candidate pose pool $\{ (\bm{\tilde{x}}_t, \bm{\tilde{y}}_t) \}$ with $T$ samples for the training sample $(\bm{x}, \bm{y})$, where $t \in \{1, 2, ..., T\}$.
PoseTrans select the rarest one $(\bm{\tilde{x}}_{t^{*}}, \bm{\tilde{y}}_{t^{*}})$ among the candidate pose pool by:
\begin{equation}
   \begin{gathered}
   t^{*} = \underset{t}{\operatorname{argmin}} \left( \sum_{n=1}^{N} \alpha_n w_t^n \right),
   \end{gathered}
\end{equation}
where $\bm{w}_t = \{w_t^1, w_t^2, \dots, w_t^N\}$ is the predicted probability of $\bm{\tilde{y}}_t$ belonging to $N$ Gaussian components by the fitted PCM.
We consider the transformed sample $(\bm{\tilde{x}}_{t^{*}}, \bm{\tilde{y}}_{t^{*}})$ with the minimal weighted sum of components' density as the rarest and select it as a new training sample.

\section{Experiments}

\subsection{Datasets and Evaluation}

\textbf{Datasets.}
To verify the effectiveness of our proposed data augmentation approach, we conduct extensive experiments on popular datasets.
\textbf{(1)} MS-COCO \cite{lin2014microsoft} pose estimation dataset. Our models are trained on the train set only and evaluated on the val set and the test-dev set. DensePose \cite{alp2018densepose} provides a small portion of human parsing annotations for the MS-COCO dataset. To verify the performance on rare poses, both the traditional evaluation metrics (\ie AP/AR) and newly designed metrics (balanced AP/AR) are used for evaluation. The base learning rate of 1e-3, and decay the learning rate to 1e-4 and 1e-5 at the 170-th and 200-th epochs respectively.
\textbf{(2)} PoseTrack'18 \cite{andriluka2018posetrack} dataset. Following common settings~\cite{mmpose2020}, we pre-train the model on the MS-COCO dataset and fine-tune it on the PoseTrack'18 dataset for 20 epochs. The basic learning rate is 1e-4 and drops to 1e-5 at 10 epochs then 1e-6 at 15 epochs. We test the model on the PoseTrack'18 validation set using the ground truth bounding boxes, and evaluate the AP on the whole body and also on different parts of the human.
Due to the limited space, the results of some experiments are placed in the supplementary material.

\textbf{Evaluation metrics.}
We follow~\cite{lin2014microsoft} to use Average Precision (AP) and Average Recall (AR) for evaluation on MS-COCO~\cite{lin2014microsoft}. They are based on object keypoint similarity (OKS), which measures the distance between predicted keypoints and ground-truth keypoints normalized by the scale of the object. AP$_{50}$ (AP at OKS = 0.5), AP$_{75}$ (AP at OKS = 0.75), AP$^{M}$ for medium objects, and AP$^{L}$ for large objects are reported.

\textbf{Balanced AP/AR.}
Since existing datasets mostly suffer the long-tailed distribution problem, simply calculating the AP/AR tends to ignore the minor pose categories.
To solve this problem, we design the balanced AP/AR, which we term $\text{AP}_{\text{BAL}}$, $\text{AR}_{\text{BAL}}$. We first classify the ground-truth poses into categories based on the fitted PCM. Then we calculate the standard AP/AR separately for each category and calculate the average precision/recall among \emph{categories} instead of \emph{samples}. Therefore, $\text{AP}_{\text{BAL}}$ and $\text{AR}_{\text{BAL}}$ assign the same weights to all pose categories, which is helpful to analyze the ``unbiased'' performance.

\subsection{Implementation Details}
PoseTrans can be integrated into the training pipeline of any existing pose estimators together with other common data augmentation strategies.
Except for the small portion of images that have human parsing annotations, we leverage DensePose \cite{alp2018densepose} model for human parsing which segments humans into 14 semantic parts.
In PCM, we have $N = 20$ and cluster the poses into 20 categories.
We implement PoseTrans with scaling ($s \in [0.75, 1.25]$), rotating ($r \in [-35^{\circ}, 35^{\circ}$]), and apply it with the probability $p = 0.5$ for every limb in the training examples.
We filter out the implausible samples whose plausibility is less than $E = 0.7$ and repeat PTM until the candidate pose pool has $T = 5$ augmented samples.

\begin{table*}[!t]
  \caption{
     \textbf{Improvements} on MS-COCO \texttt{val} set and \texttt{test-dev} set.
     PoseTrans consistently boosts the performance of the state of the arts.
     }
     \begin{center}
     \scalebox{0.8}{
     \begin{tabular}{l|c|accccc|accccc}
     \hline
     \multirow{2}{*}{Method} 
       & \multirow{2}{*}{Input size}        & \multicolumn{6}{c}{MS-COCO \texttt{val}}  & \multicolumn{6}{c}{MS-COCO \texttt{test-dev}}  \\
     \cline{3-14} 
     & &  AP  & $\text{AP}^{50}$ & $\text{AP}^{75}$ &$\text{AP}^{M}$ & $\text{AP}^{L}$  & AR  &  AP  & $\text{AP}^{50}$ & $\text{AP}^{75}$  & $\text{AP}^{M}$ & $\text{AP}^{L}$ & AR   \\
     \hline
     \multicolumn{14}{c}{\textit{Bottom-up methods w/o multi-scale test}}\\
     \hline
     AE\cite{newell2017associative} + HRNet-W32\cite{sun2019deep}    &$512\times512$    &   64.4        & 86.3        &  72.0 & 57.1 & 75.6  &   71.0  & 64.1       & 86.3      &  70.4  & 57.4 & 73.9 &  70.4   \\
     + PoseTrans (Ours)   &$512\times512$    &   \textbf{66.2}   & \textbf{86.4}         & \textbf{72.1}  & \textbf{59.3} & \textbf{76.5}   &  \textbf{71.6}    &     \textbf{65.4}  &   \textbf{87.6}    & \textbf{72.1}  & \textbf{58.8} & \textbf{74.7}   & \textbf{71.0}     \\
     \hline
     HigherHRNet-W32\cite{cheng2020higherhrnet} &$512\times512$   &   67.1   &   86.2    & 73.0  & 61.5 & 76.1 &  72.3   & 66.4     & 87.5   &   72.8  & 61.2 & 74.2  &  71.4    \\
  
     + PoseTrans (Ours)  & $512 \times 512$  &   \textbf{68.4}   & \textbf{87.1}      & \textbf{74.8} & \textbf{62.7} & \textbf{77.1}  &   \textbf{72.9}   &  \textbf{67.4}    &   \textbf{88.3}  &  \textbf{73.9}  & \textbf{62.1}  & \textbf{75.1}  &  \textbf{72.2}   \\
             
     \hline
     \multicolumn{14}{c}{\textit{Bottom-up methods with multi-scale test} [$\times2$, $\times1$, $\times0.5$]}\\
  
     \hline
     AE\cite{newell2017associative} + HRNet-W32\cite{sun2019deep} &$512\times512$  & 68.5 & 87.1  & 75.1 & 64.0 & 76.8 & 73.9     & 68.1   & 88.3     &  75.1  & 63.8 & 74.9  &  72.9     \\
     
     + PoseTrans (Ours)   & $512\times512$  &   \textbf{70.5}   & \textbf{87.8}   &  \textbf{76.7}  & \textbf{65.1} &  \textbf{78.1}    & \textbf{75.2}    & \textbf{69.4}         &   \textbf{88.8}    &  \textbf{76.3} & \textbf{64.4} & \textbf{76.2}  &  \textbf{74.2}   \\
     \hline
     HigherHRNet-W32\cite{cheng2020higherhrnet} &  $512 \times 512$  & 69.9  &  87.1  &  76.0 & 65.3 & 77.0 &  74.7    & 68.8       & 88.8       &   75.7   &  64.4 & 75.0 & 73.5  \\
  
     + PoseTrans (Ours)   &$512 \times 512$   &  \textbf{71.2}    & \textbf{88.2}   & \textbf{77.2}     & \textbf{66.5} & \textbf{78.0} &   \textbf{75.3}   & \textbf{69.9}       & \textbf{89.3}            & \textbf{77.0}  & \textbf{65.2} & \textbf{76.2}  &  \textbf{74.3}   \\
     \hline
     \multicolumn{14}{c}{\textit{Top-down methods}}\\
     \hline
     SBL-ResNet-50\cite{xiao2018simple}  & $256\times 192$ &   70.4    & 88.6             & 78.3      & 67.1 & 75.9     & 76.3   & 70.2     & 90.9     &  78.3 & 67.1 & 75.9  &  75.8  \\
     + PoseTrans (Ours)  & $256\times 192$ &    \textbf{72.3}       & \textbf{89.9}       & \textbf{80.0 }   & \textbf{68.3} & \textbf{79.2}  &  \textbf{77.8}  & \textbf{71.5}       & \textbf{91.8}     &  \textbf{80.0}  & \textbf{68.1} & \textbf{77.3}  & \textbf{77.0}    \\
     \hline

     SBL-ResNet-101\cite{xiao2018simple}  & $256\times 192$  & 71.4 &  89.3 & 79.3 & 68.1 & 78.1  &  77.1  & 71.1 & 91.5 & 79.6 & 67.7 &76.8 & 76.6  \\
     + PoseTrans (Ours)  & $256\times 192$    & \textbf{72.7}  & \textbf{90.0}  &  \textbf{ 80.7} &   \textbf{69.5} & \textbf{78.8} & \textbf{78.3}    & \textbf{71.8}  & \textbf{91.6}  &  \textbf{80.3} &   \textbf{68.3} & \textbf{77.5} & \textbf{77.3}   \\
     \hline

     HRNet-W32\cite{sun2019deep} &  $256 \times 192$ &   74.4   & 90.5        & 81.9      & 70.8 & 81.0     &  79.8   & 73.5  & 92.2   & 82.0  & 70.4 & 79.0 &  79.0   \\
     + PoseTrans (Ours) & $256 \times 192$  &   \textbf{75.5}   & \textbf{91.0}   & \textbf{82.9}  & \textbf{71.8} & \textbf{82.2}  & \textbf{80.7}   &   \textbf{74.2}   & \textbf{92.4}    &  \textbf{82.5}  & \textbf{70.8} & \textbf{79.6}  &  \textbf{79.4}   \\
     \hline

     HRNet-W32\cite{sun2019deep} + Dark\cite{zhang2020distribution}&   $256 \times 192$ & 
     75.6 & 90.5 & 82.1 & 71.8 & 82.8 & 80.8  & 74.6 & 92.4 & 82.9 & 71.2 & 80.3 & 79.9  \\
     + PoseTrans (Ours)    & $256 \times 192$  & \textbf{76.0} & \textbf{90.8} & \textbf{83.0}  & \textbf{72.1} & \textbf{83.2}  & \textbf{81.1}   & \textbf{75.0} & \textbf{92.5} & \textbf{82.9}  & \textbf{71.5} & \textbf{80.6}  & \textbf{80.1}  \\
     \hline

    HRNet-W32\cite{sun2019deep}&   $384 \times 288$ &    75.8   & 90.6     & 82.7  & 71.9 & 82.8  & 80.1  &  74.9  & 92.5  & 82.8 & 71.3 & 80.9 &  80.1  \\
     + PoseTrans (Ours)   & $384 \times 288$  &    \textbf{76.5}   & \textbf{90.9}    &  \textbf{83.3}  & \textbf{72.5} & \textbf{83.3}  &  \textbf{81.5}    &   \textbf{75.4}      &   \textbf{92.5}    &  \textbf{83.0}  & \textbf{71.6} & \textbf{81.1}  &  \textbf{80.4}  \\
     \hline

     HRNet-W48\cite{sun2019deep}&   $384 \times 288$ & 76.3 &  90.8  &  82.9 & 72.3  &  83.4 & 81.2 & 75.5 & 92.5 & 83.3 & 71.9 & 81.5 & 80.5  \\
     + PoseTrans (Ours)    & $384 \times 288$  & \textbf{76.8} &  \textbf{91.0}  & \textbf{83.1} & \textbf{72.7}  & \textbf{83.7}  & \textbf{81.6}  & \textbf{75.7} &  \textbf{92.6}  & \textbf{83.4} & \textbf{72.0}  & \textbf{81.7}  & \textbf{80.6}   \\
     \hline
     \end{tabular}}
     \label{tab:coco}
     \end{center}
\end{table*}

\begin{table*}[t]
\setlength\tabcolsep{6pt}
  \caption{
     \textbf{Improvements} on PoseTrack2018 validation set.
     }
     \begin{center}
     \scalebox{0.8}{
     \begin{tabular}{l|c|ccccccc|a}
     \hline
  Method   & Input size   & Head & Sho. & Elb. & Wri. & Hip & Knee & Ank. & Total AP 
     \\
     \hline
     SBL-ResNet-50 \cite{xiao2018simple}  & $256\times 192$ &   86.5    & 87.5  & 82.3   & 75.6 & 79.9    & 78.6 & 74.0 & 81.0   \\
     + PoseTrans (Ours)   & $256\times 192$ &    \textbf{87.8}       & \textbf{89.3}       & \textbf{84.7}   & \textbf{77.7} & \textbf{82.3}  &  \textbf{81.6}  &  \textbf{75.4} &  \textbf{83.0} \\
     \hline
     HRNet-W32 \cite{sun2019deep}  &  $256 \times 192$ &   87.4   & 88.6    & 84.3  & 78.5 & 79.7     &  81.8 & 78.8 & 83.0    \\
     + PoseTrans (Ours) & $256 \times 192$  &   \textbf{88.6}   & \textbf{90.0}   & \textbf{86.2}  & \textbf{80.3} & \textbf{83.1}  & \textbf{84.9}  & \textbf{79.8} & \textbf{84.9}   \\
     \hline
     HRNet-W32 \cite{sun2019deep} &   $384 \times 288$ &    88.5   & 89.5     & 86.0  & 80.4 & 81.6  & 83.4 & 78.9 & 84.3  \\
     + PoseTrans (Ours)   & $384 \times 288$  &    \textbf{88.9}   & \textbf{90.3}    &  \textbf{87.4}  & \textbf{81.8} & \textbf{83.5}  &  \textbf{85.5} &  \textbf{80.6} &  \textbf{85.7}   \\
     \hline
     \end{tabular}}
     \label{tab:posetrack}
     \end{center}
\end{table*}

For \emph{bottom-up} methods, PoseTrans is applied on every instance in the image separately. The experimental settings are the same as~\cite{cheng2020higherhrnet}. We apply image-level random scaling ([$-25\%, 25\%$]), random rotation ([$-30^{\circ}$, $30^{\circ}$]), random translation ([$-40$px, $40$px]) and random flipping. The models are trained for $300$ epochs using the Adam optimizer~\cite{kingma2014adam}. The base learning rate is 1e-3, and it decreases to 1e-4 and 1e-5 at the $200$-th and $260$-th epochs respectively. For \emph{top-down} approaches, the experimental settings are the same as~\cite{sun2019deep}. We use the detected bounding boxes provided by Xiao \etal~\cite{xiao2018simple}. The detection boxes are first extended to a fixed aspect ratio (\ie height:width = 4:3) and then enlarged by a factor of $1.25$ to include some context. We apply random scaling ([$-35\%, 35\%$]), random rotation ([$-45^{\circ}$, $45^{\circ}$]), random flipping, and half-body crops. The models are trained on $16$ GPUs for $210$ epochs. We use Adam optimizer~\cite{kingma2014adam} for training.
All networks are pre-trained on the ImageNet dataset~\cite{russakovsky2015imagenet}.

\begin{table}[t]
  \centering
  \caption{
  \textbf{(a)} Improvements of Balanced AP/AR on MS-COCO \texttt{val} set. \textbf{(b)} Comparisons of data augmentation techniques on MS-COCO \texttt{val} set. HRNet-W32 with an input size of $256 \times 192$ is adopted as the baseline. Results marked with `*' are reported by~\cite{pytel2021tilting} using CascadeRCNN bounding boxes.
  } 
  \begin{subtable}[b]{0.45\textwidth}
    \centering
    \scalebox{0.7}{
     \begin{tabular}{l|c|cc|aa}
     \hline
     \multirow{2}{*}{Method}    & \multirow{2}{*}{Input size}        & \multicolumn{4}{c}{MS-COCO \texttt{val}}    \\
     \cline{3-6}
     & &  AP   & AR  & $\text{AP}_{\text{BAL}}$ & $\text{AR}_{\text{BAL}}$    \\
     \hline
     SBL-ResNet50~\cite{xiao2018simple}   & $256\times 192$ &   70.4        & 76.3 &  60.6     &  66.3      \\
     + PoseTrans (Ours)   & $256\times 192$ &    \textbf{72.3}       &  \textbf{77.8} &  \textbf{63.8}  &   \textbf{69.6}     \\
     \hline
     HRNet-W32 \cite{sun2019deep} &  $256 \times 192$ &   74.4    &  79.8  &   65.4 &   72.3     \\
     + PoseTrans (Ours) & $256 \times 192$  &   \textbf{75.5}   & \textbf{80.7} &  \textbf{67.9}  & \textbf{73.8}    \\
     \hline
     HRNet-W32 \cite{zhang2020distribution}&   $384 \times 288$ &    75.8   & 80.1 &  67.7   &  73.8     \\
     + PoseTrans (Ours)   & $384 \times 288$  &    \textbf{76.5}     &  \textbf{81.5}  &  \textbf{68.9}  & \textbf{74.2}    \\
     \hline
     \end{tabular}}
    \caption{}
    \label{tab:coco_bal}
  \end{subtable}
  \quad
  \begin{subtable}[b]{0.5\textwidth}
    \centering
    \setlength{\tabcolsep}{2.0pt}
    \scalebox{0.7}{
      \begin{tabular}{l|cccccc}
      \hline
      Method   &   AP   & $\text{AP}^{50}$ & $\text{AP}^{75}$  &AR  \\
      \hline
      Baseline~\cite{sun2019deep}  &74.4  & 90.5  & 81.9  & 79.8 \\
      + Cutout*~\cite{devries2017improved}   & 74.5  & 90.5  & 81.7    & 78.8 \\
      + GridMask~\cite{chen2020gridmask} & 74.7 & 90.6 & 82.0 & 80.1 \\
      + Photometric Distortion~\cite{bochkovskiy2020yolov4} & 74.6 & 90.3 & 81.9 & 80.0 \\
      + AdvMix~\cite{wang2021human} & 74.7 &  -    & -    & -     \\
      + InstaBoost~\cite{fang2019instaboost} &  74.7  &   90.5    & 82.0   & 80.1    \\
      + ASDA~\cite{bin2020adversarial} &  75.2  &   \textbf{91.0}    &  82.4  & 80.4    \\
      \hline
      + PoseTrans (Ours) & \textbf{75.5}  & \textbf{91.0} & \textbf{82.9}   & \textbf{80.7}     \\
      \hline
      \end{tabular}
   }
    \caption{}
    \label{tab:data-aug}
  \end{subtable}
\end{table}

\subsection{Improvement of state-of-the-art methods by PoseTrans}

\textbf{Improvement of AP/AR.}
Table \ref{tab:coco} reports the performance improvement of AP and AR on the MS-COCO \texttt{val} and MS-COCO \texttt{test-dev} set, where PoseTrans is applied to recent state-of-the-art pose estimators, \ie SBL~\cite{xiao2018simple}, HRNet~\cite{sun2019deep}, and HigherHRNet~\cite{cheng2020higherhrnet}.
Table \ref{tab:posetrack} show the performance improvement on the PoseTrack dataset.
PoseTrans consistently boosts the performance of both top-down and bottom-up approaches in various datasets.

\textbf{Improvement of $\text{AP}_{\text{BAL}}$ and $\text{AR}_{\text{BAL}}$.} The results of $\text{AP}_{\text{BAL}}$ and $\text{AR}_{\text{BAL}}$ are reported in Table \ref{tab:coco_bal}.
To calculate the new metrics, we use the bounding boxes and keypoint annotations to determine the category of predicted poses.
Thanks to the design of PCM and PTM, PoseTrans increases the diversity of rare poses and balances the distribution, which enables PoseTrans to bring more improvements on the newly proposed $\text{AP}_{\text{BAL}}$/$\text{AR}_{\text{BAL}}$ than traditional AP/AR.

\subsection{Comparisons with other data augmentation techniques}

In Table~\ref{tab:data-aug}, we compare PoseTrans with other data augmentation techniques, including non-learning \cite{devries2017improved,chen2020gridmask,bochkovskiy2020yolov4} and learning/strategy-based methods \cite{wang2021human,fang2019instaboost}.

For non-learning methods, Cutout~\cite{devries2017improved} randomly selects a rectangle region around the keypoint and fills in random values.
GridMask~\cite{chen2020gridmask} evenly replaces multiple
rectangle regions in an image with all zeros. For Photometric Distortion, we follow~\cite{bochkovskiy2020yolov4} to adjust the brightness, contrast, hue, saturation, and noise of an image. These general data techniques are proven to be effective for image classification. However, they do not bring significant improvements for human pose estimation. Similar conclusions have also been reached by previous works~\cite{pytel2021tilting}. This is probably because such techniques introduce undesirable artifacts and do not increase the diversity of human poses.

For learning/strategy-based methods, AdvMix~\cite{wang2021human} applies adversarial training to learn to mix up augmented samples generated by GridMask~\cite{chen2020gridmask} and AutoAugment~\cite{cubuk2018autoaugment}. InstaBoost~\cite{fang2019instaboost} is a recently proposed data augmentation technique which is originally designed for instance segmentation. It conducts crop-paste augmentation guided by the appearance consistency heatmaps. However, the improvements of AdvMix and InstaBoost are only marginal.
ASDA \cite{bin2020adversarial} also employs human parsing and augments images by pasting the segmented body parts.
PoseTrans outperforms all these approaches, which validates the importance of increasing the diversity of the human body poses. 

Kindly note that PoseTrans is also complementary to other techniques. Effectively combining these techniques may further improve the final performance.
As shown in the third row from the bottom in Table \ref{tab:coco}, combining PoseTrans with DarkPose \cite{zhang2020distribution} can further gain improvements.

\subsection{Ablation Studies}

\textbf{Effect of PTM.}
Without using the PTM, we perform the over-sampling~\cite{chawla2002smote} and re-weighting~\cite{elkan2001foundations} strategies, which are two popular methods to tackle the long-tailed problem.
The over-sampling method raises the frequency level of the minor categories by duplicating the long-tailed data samples during model training. The re-weighting method assigns higher loss weights to rare samples and thus increases their importance.
Based on the clustering results of PCM, we implement these methods as baselines, as shown in Table~\ref{tab:abl_ptm}. 
By increasing the importance of long-tailed training samples, both the over-sampling marginally improve the $\text{AP}_{\text{BAL}}$.
However, such approaches do not increase the diversity of the data, which leads to slight performance drops on AP and AR. 
With the design of PTM, our proposed PoseTrans creates diverse long-tailed samples, which significantly outperforms the baseline methods.

\textbf{Effect of PCM.}
Without PCM, PoseTrans randomly samples a transformed pose obtained from PTM as the training sample, instead of picking the ``rarest'' pose in the candidate pose pool. Note that, ``w/o PCM'' is 
equivalent to the case of $T=1$ in PoseTrans.
The studies of w/o PCM and the number of $T$ in PCM are shown in Table~\ref{tab:abl_pcm}.
By providing simple disturbance to training data, w/o PCM increases the generalization of the model, which leads to some performance improvements. 
While with the aid of PCM, our full model learns to alleviate the long-tailed distribution problem of the training set by selecting transformed poses, which brings greater performance gains, especially on $\text{AP}_{\text{BAL}}$/$\text{AR}_{\text{BAL}}$.
Also, a larger candidate pose pool (\ie greater $T$) leads to better performance.
However, $T$ greater than $5$ will not bring more performance boost.

\textbf{Effect of pose discriminator.}
Without the pose discriminator ($D$), some implausible poses will lead to performance degradation as shown in Table \ref{tab:abl_dis}.
Since the scale $s$ and rotation $r$ parameters are sampled from a normal distribution and are restricted to $[0.75, 1.25]$ and $[-35^{\circ}, 35^{\circ}]$ in the implementation, a majority of the randomly generated poses are plausible. In this situation, the PTM without the pose discriminator can still benefit the model.

\begin{table}[t]
  \centering
  \caption{
  \textbf{(a)} Ablation studies of PTM. The over-sampling and re-weighting methods are based on the clustering results of PCM. \textbf{(b)} Ablation studies of PCM. HRNet-W32 with the input size of $256 \times 192$ is adopted for experiments. }
  \begin{subtable}[b]{0.45\textwidth}
    \centering
    \scalebox{0.8}{
   \begin{tabular}{l|cccccc}
   \hline
   Method   &   AP   & AR & $\text{AP}_{\text{BAL}}$ & $\text{AR}_{\text{BAL}}$ \\
   \hline
   Baseline & 74.4  & 79.8  &   65.4 &   72.3 \\
   Over-sampling~\cite{chawla2002smote}  &  74.3  &  79.7  &  66.0  & 72.3 \\
   Re-weighting~\cite{elkan2001foundations}  & 74.2  & 79.6  &  65.8  & 72.2 \\
   \hline
   PoseTrans (Ours) & \textbf{75.5}   & \textbf{80.7} &  \textbf{67.9}  & \textbf{73.8}  \\
   \hline
   \end{tabular}
   }
    \caption{}
    \label{tab:abl_ptm}
  \end{subtable}
  \quad
  \begin{subtable}[b]{0.5\textwidth}
    \centering
    \setlength{\tabcolsep}{2.0pt}
    
    \scalebox{0.8}{
    \begin{tabular}{l|cccccc}
    \hline
    Method   &   AP   & AR & $\text{AP}_{\text{BAL}}$ & $\text{AR}_{\text{BAL}}$ \\
    \hline
    Baseline \cite{sun2019deep} & 74.4 & 79.8  &  65.4  & 72.3 \\
    w/o PCM & 74.9 & 80.1  & 66.1 & 72.6 \\
    PoseTrans ($T = 3$)  & 75.2  & 80.3  & 67.2  & 72.9 \\
    \hline
    PoseTrans ($T = 5$) &\textbf{75.5}   & \textbf{80.7} &  \textbf{67.9}  & \textbf{73.8}   \\
    \hline
    \end{tabular}
    }
    \caption{}
    \label{tab:abl_pcm}
  \end{subtable}
\end{table}

\begin{table}[t]
  \centering 
  \caption{
  \textbf{(a)} Ablation studies of Discriminator ($D$). \textbf{(b)} Comparison with the variants of PoseTrans.
  } 
  \begin{subtable}[b]{0.45\textwidth}
    \centering
    \scalebox{0.8}{
      \begin{tabular}{l|cccccc}
      \hline
      Method   &   AP   & AR & $\text{AP}_{\text{BAL}}$ & $\text{AR}_{\text{BAL}}$ \\
      \hline
      PoseTrans w/o $D$   & 75.0  & 80.1  & 66.5  & 72.8 \\
      \hline
      PoseTrans & \textbf{75.5}   & \textbf{80.7} &  \textbf{67.9}  & \textbf{73.8}  \\
      \hline
      \end{tabular}
      }
    \caption{}
    \label{tab:abl_dis}
  \end{subtable}
  \quad
  \begin{subtable}[b]{0.5\textwidth}
    \centering
    \setlength{\tabcolsep}{2.0pt}
    \scalebox{0.8}{
    \begin{tabular}{l|cccccc}
    \hline
    Method   &   AP   & AR & $\text{AP}_{\text{BAL}}$ & $\text{AR}_{\text{BAL}}$ \\
    \hline
    PoseTrans-Adv   & 72.7  & 78.4  & 65.2  & 71.5 \\
    \hline
    PoseTrans-Par   & 75.3  & 80.4  & 67.3  & 73.3 \\
    \hline
    PoseTrans & \textbf{75.5}   & \textbf{80.7} &  \textbf{67.9}  & \textbf{73.8}  \\
    \hline
    \end{tabular}
    }
    \caption{}
    \label{tab:abl_adv}
  \end{subtable}
\end{table}

\textbf{Comparison with the adversarial learning variant.}
Inspired by recent works \cite{wang2021human,peng2018jointly,bin2020adversarial} on adversarial data augmentation, we also build an adversarial training variant of PoseTrans, which we refer to as PoseTrans-Adv. PoseTrans-Adv has an additional generator that predicts the rotation $r$ and scale $s$ for a given single human image $\bm{x}$. During training, the generator is asked to confuse the pose estimation model by maximizing the loss of the pose estimator. However, we observe that the generator will soon learn to choose the maximum rotation and scale for every training sample, which actually decreases the diversity of the training set. This leads to performance degradation in all the evaluation metrics as shown in the first row of Table~\ref{tab:abl_adv}.

\textbf{Comparison with PoseTrans-Par on the MS-COCO dataset.}
As mentioned above, DensePose \cite{alp2018densepose} provides a small portion of human parsing annotations for the MS-COCO dataset. 
Here, we compare with the PoseTrans-Par variant that replaces the human annotations with the pseudo-labels obtained from the parsing model. As shown in the second row of Table \ref{tab:abl_adv}, without human annotations, the performance of PoseTrans-Par is comparable with PoseTrans.

\subsection{Analysis}

\begin{figure*}[h]
  \centering
  \includegraphics[width=0.98\linewidth]{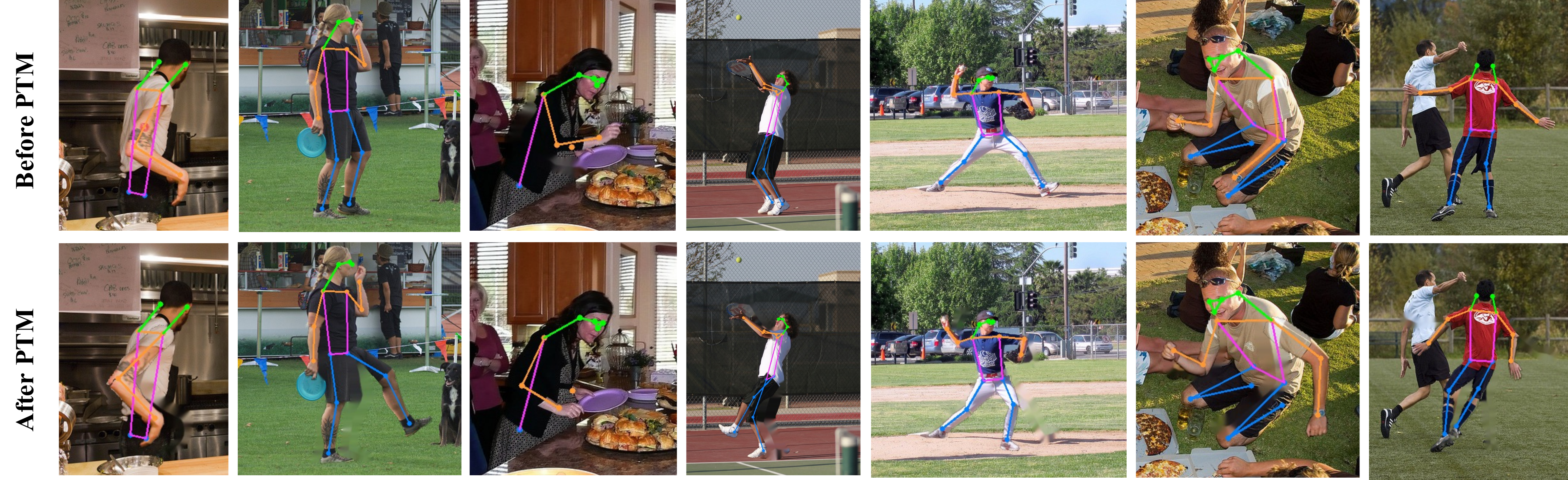}
  \caption{
     \textbf{Visualizations} of PoseTrans augmented samples. We observe that our proposed method generates more diverse body postures which facilitates the model training and improves its generalization ability.
  }
  \label{vis_ptm}
\end{figure*}

\textbf{Visualizations of the augmented samples.}
In Fig. \ref{vis_ptm}, we visualize the original image and the augmented sample by PoseTrans. It can be observed that our proposed method generates diverse and plausible body postures that facilitate the model training and improve its generalization ability.

\begin{figure*}[t]
  \centering
  \includegraphics[width=0.98\linewidth]{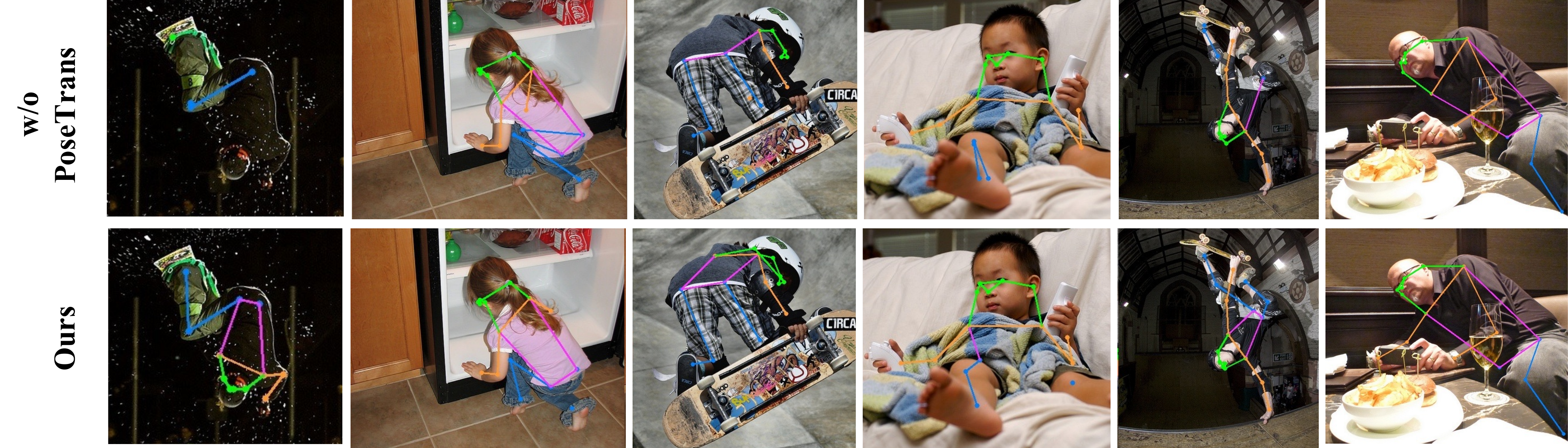}
  \caption{
     \textbf{Qualitative comparisons} of vanilla HRNet \cite{sun2019deep} (upper row) and HRNet trained with PoseTrans (bottom row).
     PoseTrans improves the human pose estimation results, especially for rare poses.
  }
  \label{vis_comp}
\end{figure*}

\textbf{Visualizations of pose estimation results.}
In Fig.~\ref{vis_comp}, we visualize pose estimation results obtained by HRNet \cite{sun2019deep}.
We observe that vanilla HRNet is easily confused by infrequent and difficult poses, \eg upside-down postures and serious occlusions.
By generating training samples with diverse rare poses, our PoseTrans improves the performance in these challenging cases.

\textbf{Limitations.}
Our limitations mainly lie in the artifacts produced by the inpainting method and the accuracy of the human parsing model. We choose a simple non-data-driven inpainting method in pose transformation for efficiency.
An advanced inpainting and parsing model with higher resolution inputs may bring more improvements in pose estimation.

\section{Conclusions}

In this paper, we study the performance degradation caused by unbalanced data distribution on human pose estimation.
To tackle this issue, we propose PoseTrans with PTM, PCM, and a pose discriminator to create diverse and plausible training samples that have infrequent poses.
Comprehensive experiments on public benchmarks demonstrate the effectiveness of our method, especially on rare poses. 
Our implementation of PoseTrans is simple and efficient, which can be easily integrated into the training pipeline of existing pose estimators.
We hope our work will draw the community's attention to the long-tail problem in human pose estimation and provide inspiration on how to tackle it for other tasks.

\textbf{Acknowledgement.}
This work is supported in part by the National Natural Science Foundation of China under Grant 62122010 and Grant 61876177, in part by the Fundamental Research Funds for the Central Universities, and in part by the Key Research and Development Program of Zhejiang Province under Grant 2022C01082. 
Ping Luo is supported by the General Research Fund of HK No.27208720, No.17212120, and No.17200622.

%%%%%%%%% REFERENCES

% \clearpage\mbox{}Page \thepage\ of the manuscript.
% \clearpage\mbox{}Page \thepage\ of the manuscript.

% This is the last page of the manuscript.
% Now we have reached the maximum size of the ECCV 2022 submission (excluding references).
% References should start immediately after the main text, but can continue on p.15 if needed.

\clearpage
\bibliographystyle{splncs04}
\bibliography{egbib}
\end{document}